\documentclass[12pt]{l4dc2020} 

\title[Risk Averse DRL]{Improving Robustness via Risk Averse Distributional Reinforcement Learning}

\usepackage{booktabs}       
\usepackage{amsfonts}       
\usepackage{nicefrac}       
\usepackage{microtype}      

\usepackage{algorithm}
\usepackage{algorithmic}
\usepackage{amsmath}
\usepackage{verbatim}
\usepackage{graphicx}
\usepackage{xcolor}
\newcommand{\cA}{{\mathcal A}}
\newcommand{\cX}{{\mathcal X}}
\newcommand{\mE}{{\mathbb E}}

\newcommand{\chen}{\color{cyan}}

\author{%
 \Name{Rahul Singh}   \Email{rasingh@gatech.edu}\\
 \addr School of Aerospace Engineering, Georgia Institute of Technology, Atlanta, GA 
 \AND
 \Name{Qinsheng Zhang} \thanks{Equal Contribution} \Email{qzhang419@gatech.edu}\\
 \addr School of Aerospace Engineering, Georgia Institute of Technology, Atlanta, GA 
 \AND
 \Name{Yongxin Chen} \Email{yongchen@gatech.edu}\\
 \addr School of Aerospace Engineering, Georgia Institute of Technology, Atlanta, GA 
}

\begin{document}

\maketitle

\begin{abstract}%
One major obstacle that precludes the success of reinforcement learning in real-world applications is the lack of robustness, either to model uncertainties or external disturbances, of the trained policies. Robustness is critical when the policies are trained in simulations instead of real world environment. In this work, we propose a risk-aware algorithm to learn robust policies in order to bridge the gap between simulation training and real-world implementation. Our algorithm is based on recently discovered distributional RL framework. We incorporate CVaR risk measure in sample based distributional policy gradients (SDPG) for learning risk-averse policies to achieve robustness against a range of system disturbances. We validate the robustness of risk-aware SDPG on multiple environments. 
\end{abstract}

\begin{keywords}%
Risk sensitive control, reinforcement learning, distributional reinforcement learning, robust reinforcement learning%
\end{keywords}

\section{Introduction}

Reinforcement learning (RL) has been successful in achieving human level or even better performance~\citep{MniKavDil15,SilSchSim17} in multiple games such as Atari and Go. However, one of the major factors hindering the application of RL to real-world continuous control tasks is the modeling gap between simulation and real-world which can lead to unpredictable, and often unwanted, results \citep{PinDavSukGup17}. More specifically, learning policies requires a large amount of training data, which is expensive to collect if trained directly in real-world environment. Thus, simulators are often used for learning policies before deploying to real-world problems. However, such simulation models usually contain uncertainties, i.e., a reality gap, which makes the policies trained in simulation less desirable in real applications. In this paper, we propose an algorithm to improve the robustness of RL against such model uncertainties. 


There are two popular approaches to robust RL: by minimizing the expected loss in the worst case via minimax formulations~\citep{Heg94,NilGha05,TamManXu14,PinDavSukGup17} and by considering risk-sensitive optimization criteria during training~\citep{ChoGha14,TamGlaMan15}. The relation between the two has been studied in~\cite{Jac73, CorMar99, GloDoy88,FleMcE92,FleMcE95}. Our proposed algorithm falls into the latter. In particular, we incorperate risk-sensitive criteria within distributional reinforcement learning (DRL) framework. Instead of modeling the value function as the expected sum of rewards, the DRL~\citep{BelDabMun17} framework suggests to work with the full distribution of random returns, known as value or return distribution, i.e., $Q^\pi(x,a) = \mE Z^\pi(x,a)$, where $Z^\pi(x,a)$ denotes the return distribution. The return distribution $Z$ in DRL framework provides tremendous flexibility to incorporate risk in the training process. 

In DRL, the return distribution is usually represented by discrete categorical form~\citep{BelDabMun17,MarHofBud18,QuManXu18}, quantile function~\citep{DabRowBel18,ZhaMarYao19}, or samples~\citep{FreShiMei19,SinChe19}. D4PG~\citep{MarHofBud18} and SDPG~\citep{SinChe19} are actor-critic type policy gradient algorithms based on DRL and have demonstrated much better performance~\citep{MarHofBud18,Tas18} as compared to its non-distributional counterpart (DDPG) \citep{SilLevHeeRie14} for continuous control tasks. In SDPG, the return distribution is represented via samples as opposed to discrete categorical representation in D4PG, which has shown advantages in terms of sample efficiency as well as maximum rewards. Even though D4PG and SDPG learn the return distribution, they optimize the mean value of the returns and therefore, are susceptible to model uncertainties. In this work, we incorporate risk-sensitive criteria for optimizing the policy in SDPG algorithm to achieve robustness against a range of disturbances in the system. Specifically, we focus on conditional value at risk (CVaR)~\citep{ChoGha14,ChoTamMan15}, a widely adopted risk measure. 

We perform multiple experiments to illustrate the robustness of the learned policy incorporating the risk during training against the risk-neutral policy. We demonstrate the robustness of our risk-averse SDPG algorithm against system disturbances on multiple OpenAI Gym~\citep{BroChePet16} environments for continuous control tasks including BipedalWalker, HalfCheetah, and Walker2d.

\textbf{Related Work:} There have been a few methods which accounted for risk within DRL framework including~\cite{MorSugKas10a,MorSugKas10b,DabOstSilMun18,TanZhaSal19}. The distributional-SARSA-with-CVaR proposed in ~\cite{MorSugKas10a,MorSugKas10b} deals with discrete action space with only a small number of states. Although implicit quantile network (IQN) proposed in~\cite{DabOstSilMun18} improved upon traditional deep Q-networks (DQNs)~\citep{MniKavSil15} and studied risk-sensitive policies in Atari games, it is a value function based approach and thus not suitable for tasks with continuous action space. Worst cases policy gradients (WCPG) proposed in~\cite{TanZhaSal19} models the return distribution $Z$ as Gaussian in order to calculate CVaR in closed form, but this restriction may undermine the advantages of DRL. In terms of taxonomy presented by~\cite{GarFer15}, our approach lies in the risk-sensitive criterion.

The contributions of this work are as follows: (a) We propose a novel RL algorithm to learn robust policies for continuous control tasks. Our algorithm is based on the recently discovered DRL framework. The fact that DRL learns the distribution instead of the mean of the cost-to-go function makes it suitable for risk-sensitive learning. We further took advantage our recent algorithm SDPG~\citep{SinChe19} to evaluate the risk criteria efficiently using samples. (b) We empirically evaluate the performance of our algorithm on multiple OpenAI Gym environments.

Rest of the document is organized as follows. First we briefly discuss background on DRL, SDPG, and risk measures in Section~\ref{sec:background}. Next, we present our risk-averse SDPG algorithm in Section~\ref{sec_riskaverse_sdpg} followed by experimental results in Section~\ref{sec_experiments}. Finally, Section~\ref{sec_conclusion} concludes the paper. 

\section{Background}
\label{sec:background}
We consider standard RL setting where the interaction of an agent with an environment is modeled as $(\cX, \cA, R, P, \gamma)$. Here, $\cX, \cA$ denote the state and action spaces respectively, $P(\cdot \mid x, a)$ is the transition kernel, $\gamma \in [0,1]$ is the discount factor, and $R(x,a)$ is the reward of taking action $a$ at state $x$. Our focus in this paper is on continuous state and action spaces and deterministic policies $a_t=\pi(x_t)$. Traditional RL aims to find a stationary policy $\pi$ that maximizes the Q-function which is the expected long-term discounted reward 
\begin{equation}
		Q^\pi(x,a) = \mE \left[\sum_{t=0}^\infty \gamma^t R(x_t, a_t)\right],~~ x_t \sim P(\cdot\mid x_{t-1}, a_{t-1}), a_t = \pi(x_t), x_0 = x, a_0 = a.
\end{equation}
The Q-function is characterized by Bellman's equation \citep{Bel66}
	\begin{equation}
		Q^\pi(x,a) = \mE R(x, a) + \gamma \mE Q^\pi (x', \pi(x') \mid x, a). 
	\end{equation}

\subsection{DRL}
Distributional reinforcement learning (DRL) models intrinsic randomness of return in form of full return distribution for each state-action pair
\begin{equation}
		Z^\pi(x,a) = \sum_{t=0}^\infty \gamma^t R(x_t, a_t),~~ x_t \sim P(\cdot\mid x_{t-1}, a_{t-1}), a_t = \pi(x_t), x_0 = x, a_0 = a.
\end{equation}
Apparently, Q-function is the mean of return distribution, i.e., $Q^\pi(x,a) = \mE Z^\pi(x,a)$.
The return distribution satisfies distributional Bellman's equation \citep{BelDabMun17}
	\begin{equation}\label{eq:Bellman}
		Z^\pi(x,a) = R(x,a) + \gamma Z^\pi (x', \pi(x') \mid x, a),
	\end{equation}
where the equality is in the probability sense. 

Different methods have been proposed to parameterize a return distribution in DRL. C51~\citep{BelDabMun17} and D4PG~\citep{MarHofBud18} use discrete categorical distribution, QR-DQN~\citep{DabRowBel18} and IQN~\citep{DabOstSilMun18} utilize quantiles, and VDGL~\citep{FreShiMei19} and SDPG~\citep{SinChe19} use samples to model a return distribution. These DRL algorithms have shown significant performance improvements over non-distributional counterparts in multiple environments including Atari games and DeepMind Control Suite~\citep{Tas18}. 


\subsection{SDPG}
\label{subsec:SDPG}

\begin{figure*}[h]
\centering
\vspace{-0.4cm}
\includegraphics[scale=0.60]{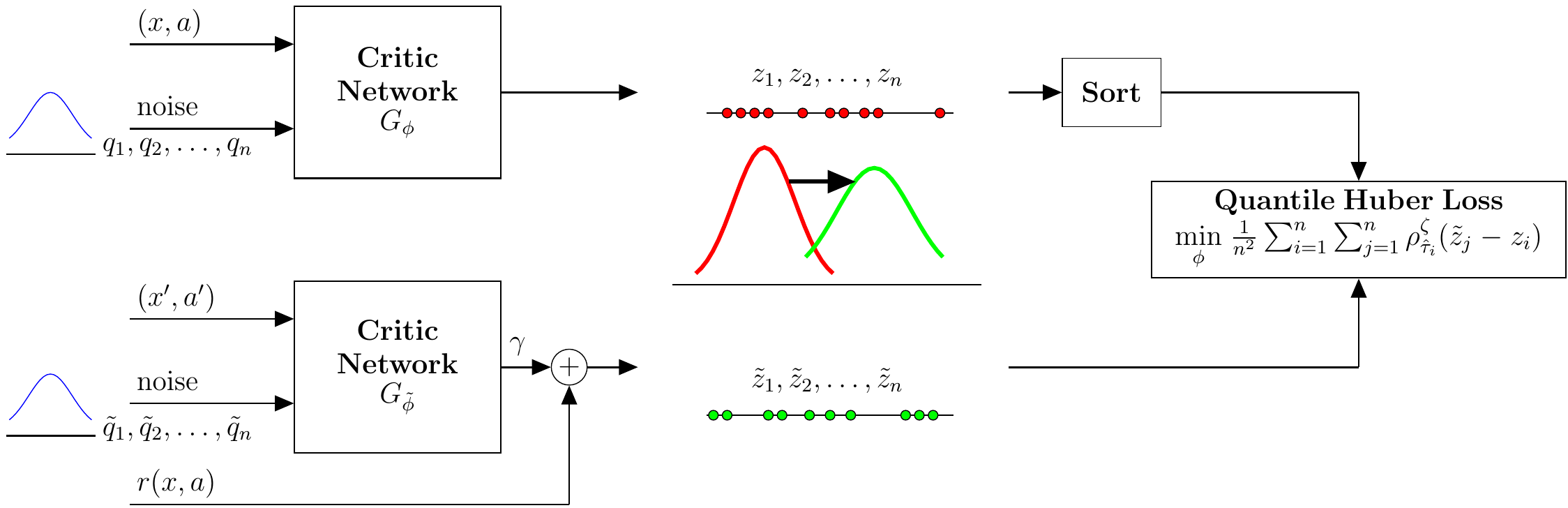}
\vspace{-0.4cm}
\caption{Flow diagram of SDPG.}
\label{fig_diagram}
\end{figure*}

Sample based policy gradient (SDPG)~\citep{SinChe19} is an actor-critic type policy gradient method within DRL framework where return distribution is represented by samples through a reparametrization technique \citep{KinWel13}. The actor network in SDPG parameterizes the policy while the critic network is trained to mimic the return distribution determined via distributional Bellman equation based on samples. A flow diagram of the critic in SDPG is shown in Figure~\ref{fig_diagram}. The critic network $G_\phi$ in SDPG learns the return distribution by utilizing quantile Huber loss~\citep{Huber1964,DabRowBel18} as a surrogate of Wasserstein distance:
\begin{equation}\label{eq:Loss_Critic}
    L_{critic}(\phi) = \mathbb{E} \left[\frac{1}{n^2} \sum_{i=1}^{n} \sum_{j=1}^{n} \rho_{\hat{\tau}_i}^{\zeta} (\tilde{z}_{j} - z_{i} ) \right],
\end{equation}
where $z_1 \geq z_2 \geq \hdots z_n$ are samples after sorting. Moreover, $\rho^{\zeta}_{\hat{\tau}_i} (v) = |\hat{\tau}_i - \delta_{\{v<0\}} | L_\zeta(v)$,
\begin{equation*}
    L_\zeta(v) = \begin{cases} 
    0.5\,v^2 & \text{if~} |v| < \zeta \\
    \zeta(|v| - 0.5~ \zeta) & \text{otherwise,}
    \end{cases}
\end{equation*}
and $\hat{\tau}_i = \frac{1}{2} (\tau_i + \tau_{i-1}),~~ i= 1,2,\ldots, n$
with $\tau_i = \frac{i}{n}$.

Using the distributional policy gradient theorem \citep{MarHofBud18}, the gradient of the loss function of the actor network $\pi_\theta$ is computed as
\begin{equation}
     \nabla_\theta L_{actor}(\theta)\!\! =\!\! \mathbb{E} \left[ \nabla_\theta \pi_\theta (x)~ \frac{1}{n} \sum_{j=1}^{n} \left[ \nabla_a z_j \right]|_{a = \pi_\theta (x)} \right].
\end{equation}

\subsection{Risk Measures}
\label{subsec:risk_measures}

The notion of risk in RL is related to the fact that even an optimal policy may perform poorly in some cases due to the stochastic nature of the problem. Risk-aware methods in RL have considered different forms of risk \citep{PraFu18} including the variance of the return, worst outcomes, exponential utility function, value at risk (VaR), and conditional value at risk (CVaR)~\citep{ChoGha14}. In this work, we focus on  CVaR.

CVaR of a random variable $Z$ at level $\alpha \in [0,1]$ is defined as\footnote{In this case we are incorporating CVaR while maximizing reward, which is opposite to incorporating CVaR while minimizing cost. Moreover, CVaR given by \eqref{eq:CVaR} is lower-tail CVaR~\citep{MorSugKas10a} resulting in risk-averse policy.}
\begin{equation}\label{eq:CVaR}
    \text{CVaR}_\alpha(Z) := \mathbb{E}~[Z~|~Z  \leq \text{VaR}_\alpha(Z)]
\end{equation}
Let $\{ z_i\}_{i=1}^{n}$ be the i.i.d. samples from the distribution of $Z$ and let $\{ z_{[i]}\}_{i=1}^{n}$ be its order statistics with $z_{[1]} \leq z_{[2]} \leq \ldots \leq z_{[n]}$. Then CVaR at level $\alpha$ can be estimated as~\citep{KolPraBha18}

\begin{equation}\label{eq:sample_cvar}
    \hat{c}_{n,\alpha} = \frac{1}{n(1-\alpha)} \sum_{i=1}^n z_i~ \mathbb{I} \{ z_i \leq \hat{v}_{n, \alpha} \},
\end{equation}
where $\mathbb{I}\{.\}$ is the indicator function, and $\hat{v}_{n,\alpha}$ is estimated VaR from samples $\hat{v}_{n, \alpha}  = z_{[\lfloor n(1- \alpha) \rfloor]}$ with $\lfloor . \rfloor$ being floor function. When $\alpha=0$, CVaR becomes expectation of the random variable which reduces to risk-neutral setting.

\section{Risk averse SDPG}
\label{sec_riskaverse_sdpg}

In order to take risk into account in policy learning, we utilize the return distribution to incorporate risk. We use CVaR as the risk-measure to learn the policy. Similar to SDPG, the risk-sensitive SDPG consists of two neural networks: a critic and an actor. The critic network $G_\phi$, parameterized by $\phi$, generates samples representing the return distribution by reparameterizing noise for each state-action pair. These samples are compared against the target samples determined via distributional Bellman equation given by~\eqref{eq:Bellman}. The quantile Huber loss is used for updating the critic network as given by Equation~\eqref{eq:Loss_Critic}.
\begin{algorithm*}[h]
   \caption{Risk-averse SDPG}
   \label{alg_risk_sdpg}
\begin{algorithmic}
   \STATE {\bfseries Require:} Learning rates $\beta_1$ and $\beta_2$, CVaR level $\alpha$, batch size $M$, sample size $n$, exploration constant $\delta$, 
   \STATE Initialize the the actor network ($\pi$) parameters $\theta$, critic network ($G$) parameters $\phi$ randomly
   \vspace{-0.3cm}
   \STATE Initialize target networks $(\tilde{\theta}, \tilde{\phi}) \leftarrow (\theta , \phi)$
   \FOR{the number of environment steps}
   \STATE Sample $M$ number of transitions $\{(x_t^{i}, a_t^{i}, r_t^{i}, x_{t+1}^{i})\}_{i=1}^M$ from the replay pool
   \STATE Sample noise $\{ q_j^i \}_{j=1}^{n} \sim \mathcal{N}(0,1) $ and $\{ \tilde{q_j}^i \}_{j=1}^{n} \sim  \mathcal{N}(0,1) $,~~for~~$i=1,\hdots,M$
   \STATE Apply Bellman update to create samples (of return distribution)
   \begin{align*}
       \tilde{z}_{j}^{i} = r_t^i + \gamma G_{\tilde{\phi}}(\tilde{q_j}^i| (x_{t+1}^i,\pi_{\tilde{\theta}}(x_{t+1}^i))) \quad \mathrm{for}~j=1,2,\hdots, n
   \end{align*}
   \vspace{-0.6cm}
   \STATE Generate samples $z_j^i = G_{\phi}(q_j^i| (x_{t}^i,a_{t}^i)) \quad \mathrm{for}~j=1,2,\hdots, n $
   \STATE Sort the samples $z^i$ in ascending order
   \STATE Update $G_{\phi}$ by stochastic gradient descent with learning rate $\beta_1$: 
   \begin{align*}
       \frac{1}{M} \sum_{i=1}^{M} \frac{1}{n^2} \sum_{j=1}^{n} \sum_{k=1}^{n} \rho_{\hat{\tau}_j}^{\zeta} (\tilde{z}_{k}^{i} - z_{j}^{i} )
   \end{align*}

   \STATE Update $\pi_{\theta}$ by stochastic gradient ascent with learning rate $\beta_2$: 
   \begin{align*}
   \frac{1}{M} \sum_{i=1}^{M} \pi_\theta (x_t^i)~  \nabla_a \left[ \frac{1}{n (1-\alpha)} \sum_{j=1}^{n} z_j^i ~ \mathbb{I} \{ z_j^i \leq  z_{\lfloor n (1- \alpha)  \rfloor}^i \} \right]_{a = \pi_\theta (x_t^i)}
   \end{align*}
   \vspace{-0.6cm}
   \STATE Update target networks $(\tilde{\theta}, \tilde{\phi}) \leftarrow (\theta , \phi)$
   \ENDFOR
   \STATE {\bfseries Actor}
   
   \hrulefill
   \REPEAT
   \STATE Observe $(x_t,a_t,x_{t+1})$ and draw reward $r_t$
   \STATE Sample action $a_{t+1} = \pi_\theta(x_{t+1}) + \delta \mathcal{N}(0,1)$
   \STATE Store $(x_t,a_t,r_t,x_{t+1},a_{t+1})$ in replay pool
   \UNTIL{learner finishes}

\end{algorithmic}
\end{algorithm*}

The actor network $\pi_\theta$, parameterized by $\theta$, outputs the action $\pi_\theta(x)$ given a state $x$. The actor network incorporates risk as feedback from the critic network $G_{\phi}$ in terms of the gradients of the empirical CVaR (given by Equation~\ref{eq:sample_cvar}) of the return distribution with respect to the actions determined by the policy. This feedback is used to update the actor network by applying distributional form of the policy gradient theorem. Therefore, the gradient of the actor network loss function is 

\begin{equation}
    \hspace{-0.2cm}\nabla_\theta L_{actor}(\theta)\!\! =\!\! \mathbb{E} \left[ \nabla_\theta \pi_\theta (x)~ \nabla_a \left[ \frac{1}{n (1-\alpha)} \sum_{j=1}^{n} z_j~ \mathbb{I} \{ z_j \leq \hat{v}_{n, \alpha} \} \right]_{a = \pi_\theta (x)} \right],
\end{equation}
where $\hat{v}_{n, \alpha} = z_{\lfloor n (1- \alpha)  \rfloor}$.

All the steps of risk-averse SDPG algorithm are described in Algorithm~\ref{alg_risk_sdpg}. The network parameters of actor and critic networks are updated alternatively in stochastic gradient ascent/descent fashion. 
\
\section{Experiments}
\label{sec_experiments}


We evaluate the robustness of our risk-averse SDPG algorithm against system disturbances on multiple OpenAI Gym~\citep{BroChePet16} continuous control tasks. For both actor and critic networks, we use a two layer feedforward neural network with hidden layer sizes of 400 and 300, respectively, and rectified linear units (ReLU) between each hidden layer. We also used batch normalization on all the layers of both networks. Moreover, the output of the actor network is passed through a hyperbolic tangent (Tanh) activation unit. In all experiments we use learning rates of $\beta_1 = \beta_2 = 1 \times 10^{-4}$, batch size $M=256$, exploration constant $\delta = 0.3$, and $\zeta = 1$. Across all the tasks, we use $n =51$ number of samples to represent return distributions. Moreover, we run each task for a maximum of 1000 steps per episode. We consider four different levels of disturbances on action forces to evaluate the robustness of learned policies at different $\alpha$ levels of CVaR. We parameterize the disturbances in terms of Gaussian noise added to action forces during evaluation. We consider the disturbances at multiple noise scales to illustrate the robustness. For each environment, we consider different $NoiseLevel$ of the disturbances depending on the highest action value corresponding to the domain. Specifically, $NoiseLevel = 0.3 \times a_{max}$ is the variance of the added zero mean Gaussian noise with $a_{max}$ being the highest possible action value corresponding to the environment. Due to the lack of a perfect actuator, the experiments model scenarios when we deploy the policy to the real-world.

\begin{figure*}[tbp]
\centering
        \begin{subfigure}
				\centering
                 \includegraphics[scale=0.18]{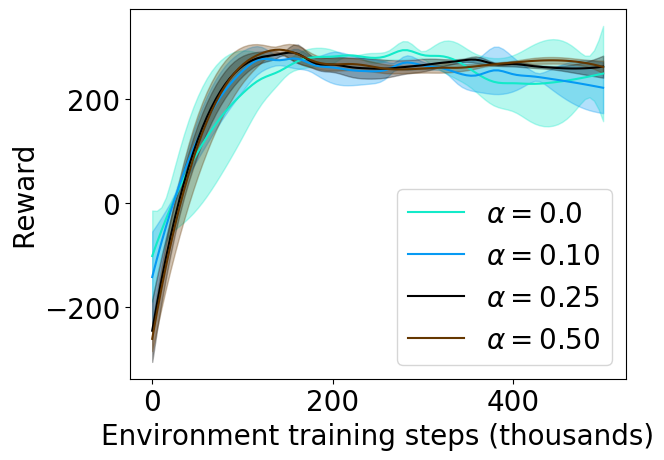}
        \end{subfigure}
        \begin{subfigure}
				\centering
                \includegraphics[scale=0.18]{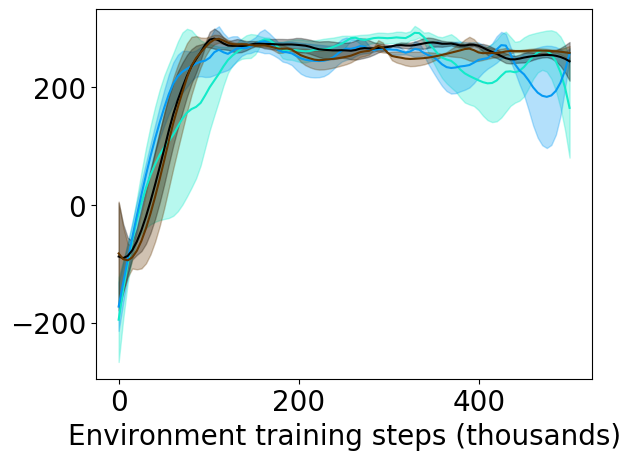}
        \end{subfigure}
        \begin{subfigure}
				\centering
                \includegraphics[scale=0.18]{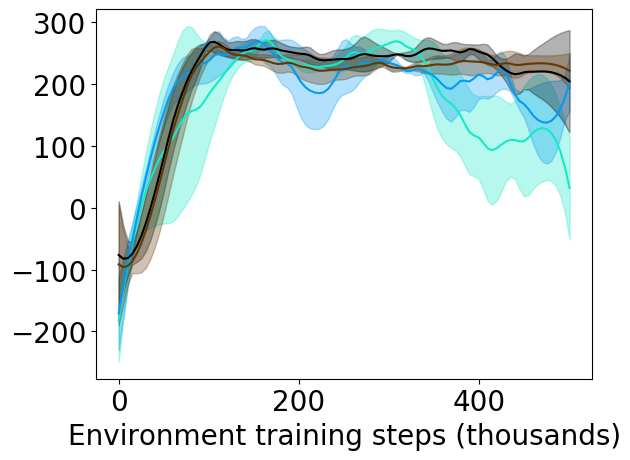}
        \end{subfigure}
        \begin{subfigure}
				\centering
                \includegraphics[scale=0.18]{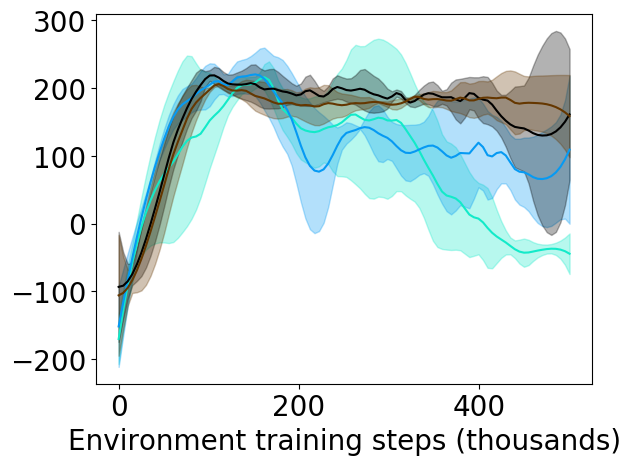}
        \end{subfigure} \\
        \begin{subfigure}
            \centering
            \includegraphics[scale=0.20]{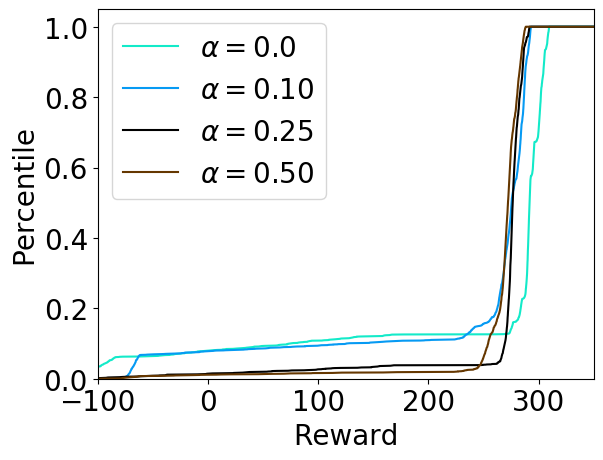}
    \end{subfigure}
    \begin{subfigure}
            \centering
            \includegraphics[scale=0.20]{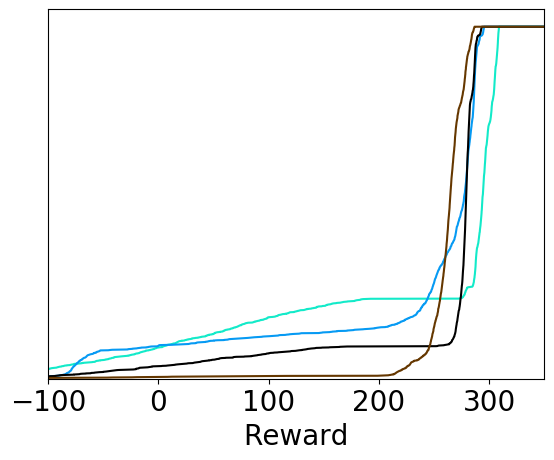}
    \end{subfigure}
    \begin{subfigure}
            \centering
            \includegraphics[scale=0.20]{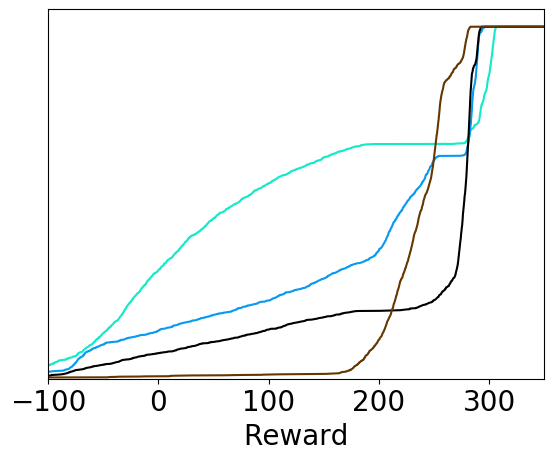}
    \end{subfigure}
    \begin{subfigure}
            \centering
            \includegraphics[scale=0.20]{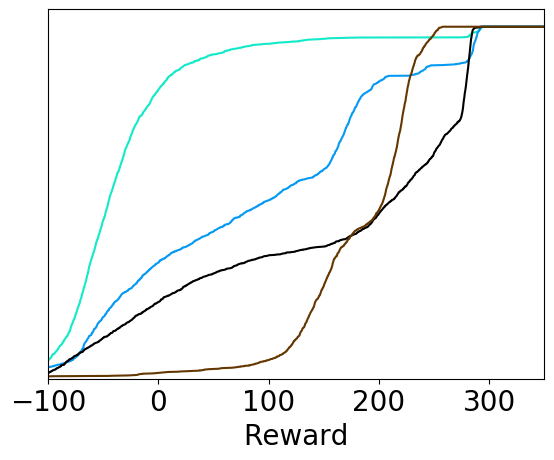}
    \end{subfigure}    
    \vspace{-0.4cm}
        \caption{BipedalWalker-v2. Top row depicts evaluation curves and the bottom row depicts the CDFs at different noise levels. The evaluations are done every 5000 environment steps in each trial over 1000 episodes. The shaded region represents standard deviation of the average returns over 5 random seeds.
        The first column is noise-free, $NoiseLevel=0$. The second column is corresponding to $NoiseLevel=0.5$. The third one is $NoiseLevel=1.0$. The final column is $NoiseLevel=1.5$. }
\label{fig:BipedalWalker-v2}
\end{figure*}

\begin{figure*}[tbp]
    \centering
            \begin{subfigure}
                    \centering
                    \includegraphics[scale=0.18]{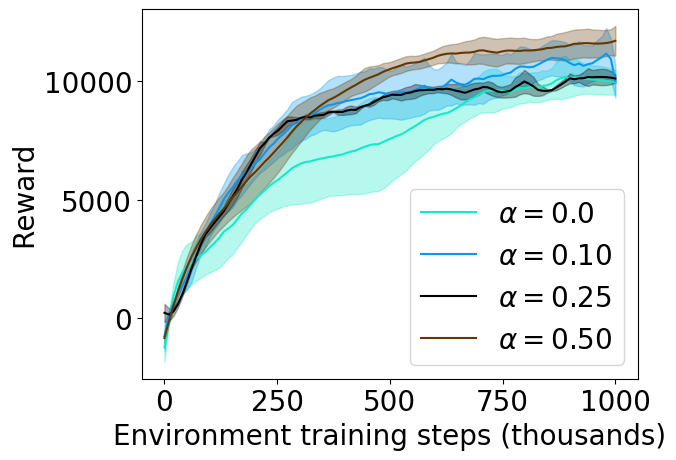}
            \end{subfigure}
            \begin{subfigure}
                    \centering
                    \includegraphics[scale=0.18]{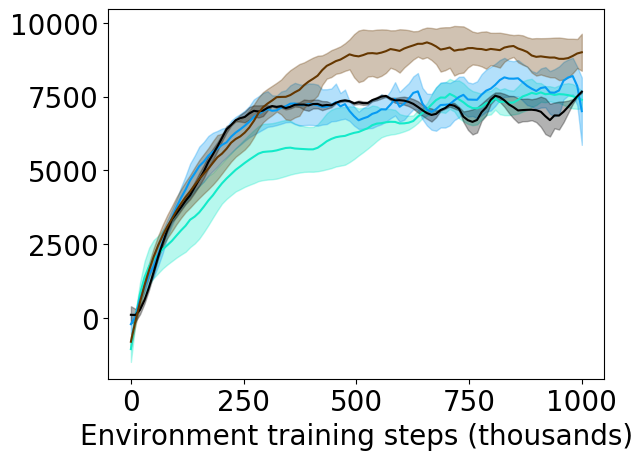}
            \end{subfigure}
            \begin{subfigure}
                    \centering
                    \includegraphics[scale=0.18]{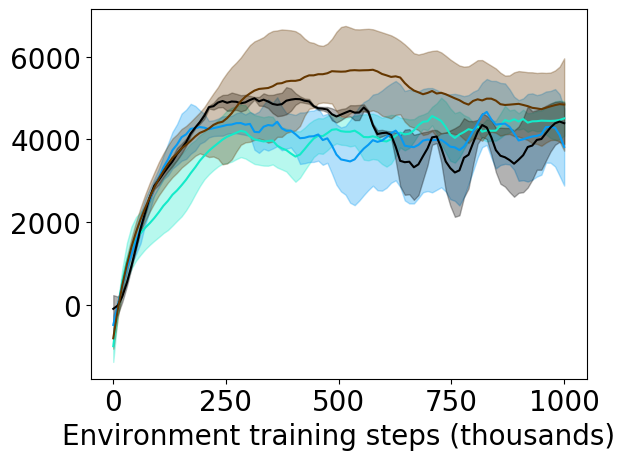}
            \end{subfigure}
            \begin{subfigure}
                    \centering
                    \includegraphics[scale=0.18]{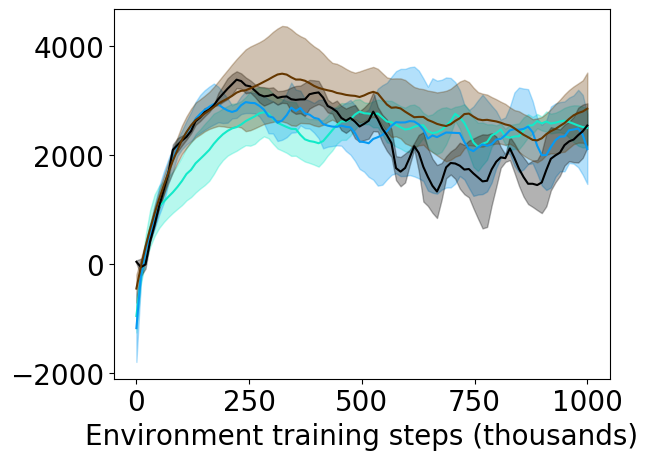}
            \end{subfigure} \\
            \begin{subfigure}
                \centering
                \includegraphics[scale=0.20]{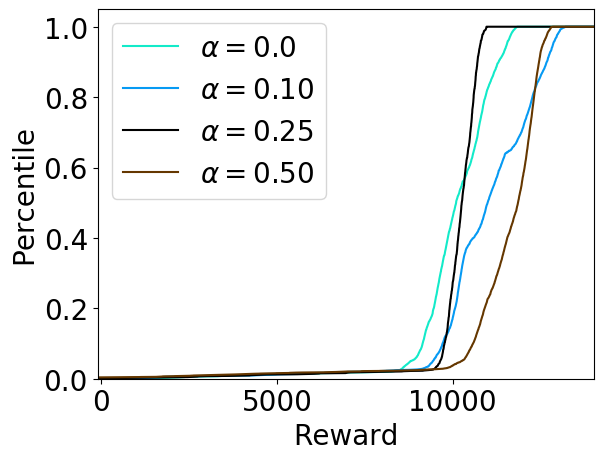}
        \end{subfigure}
        \begin{subfigure}
                \centering
                \includegraphics[scale=0.20]{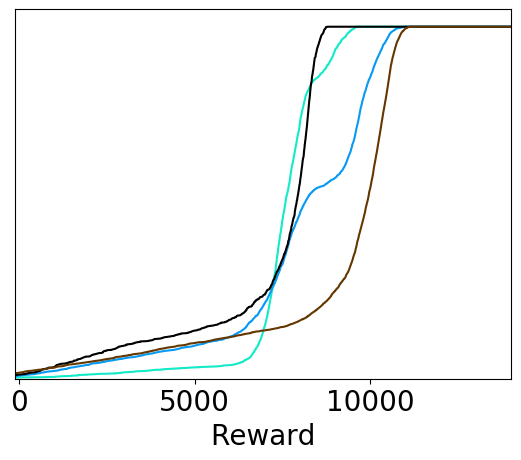}
        \end{subfigure}
        \begin{subfigure}
                \centering
                \includegraphics[scale=0.20]{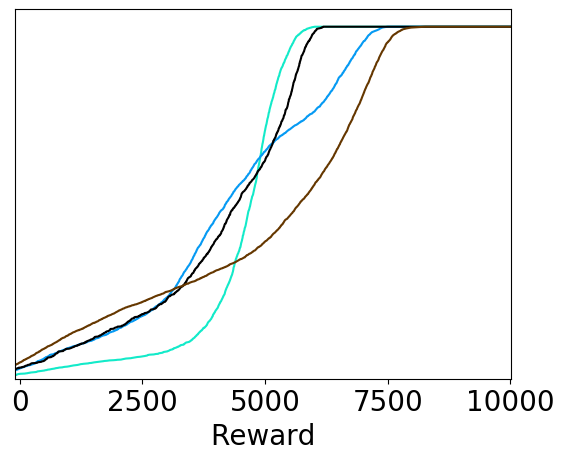}
        \end{subfigure}
        \begin{subfigure}
                \centering
                \includegraphics[scale=0.20]{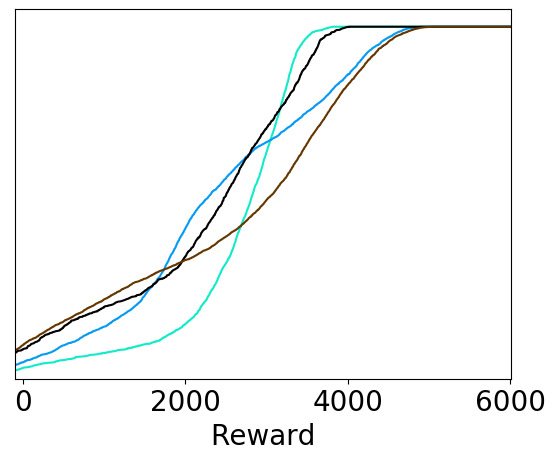}
        \end{subfigure}        
        \vspace{-0.3cm}
            \caption{HalfCheetah-v2}
\label{fig:HalfCheetah-v2}
\end{figure*}

We consider the following environments in our experiments: BipedalWalker-v2, HalfCheetah-v2, and Walker2d-v2. The task of an agent in all the three domains is to walk (run) as fast as possible without falling down and the reward is given for moving forward. We choose these environments because the reward has a large penalty when the robot falls down. These environments are not safe as compared to the other environments; the risky environment will have a value distribution with higher variance, which means there will be a higher probability that worst case scenario happens regardless of the expected reward. The state in BipedalWalker-v2 domain is 24-dimensional representing hull angle speed, angular velocity, horizontal speed, vertical speed, position of joints and joints angular speed, legs contact with ground, and $10$ lidar rangefinder measurements. The action space consists of actuator motor torques at 4 different joints. For both HalfCheetah-v2 and Walker2d-v2 domains, the state space is 18 dimensional consisting of positions, angles, and velocities of different joints while the dimension of action space is 6 consisting of actuator torques.

In each environment, we learn policies at different CVaR values $\alpha$ and evaluate the learned policies over 1000 trajectories for multiple levels of action disturbances. We also present the estimates of cumulative distribution functions (CDFs) from a total of 5000 rewards of trajectories. For all experiments in various environments, our risk-averse algorithms show similar performance during training compared to risk-neutral one. 

\begin{figure*}[tbp]
    \centering
            \begin{subfigure}
                    \centering
                    \includegraphics[scale=0.18]{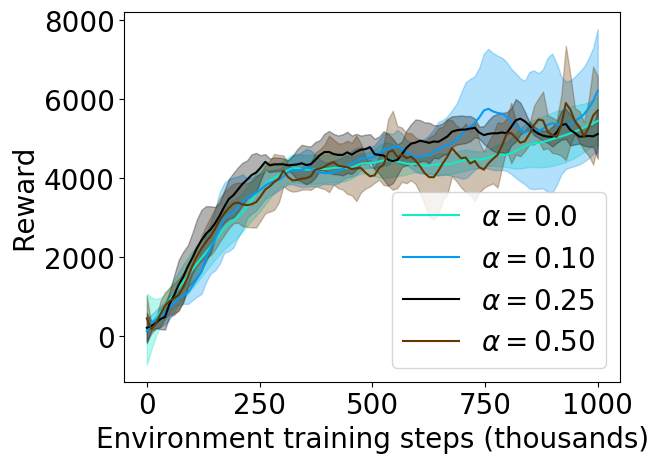}
            \end{subfigure}
            \begin{subfigure}
                    \centering
                    \includegraphics[scale=0.18]{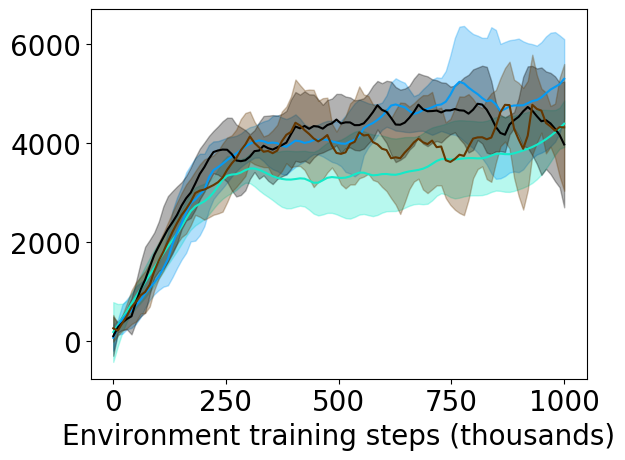}
            \end{subfigure}
            \begin{subfigure}
                    \centering
                    \includegraphics[scale=0.18]{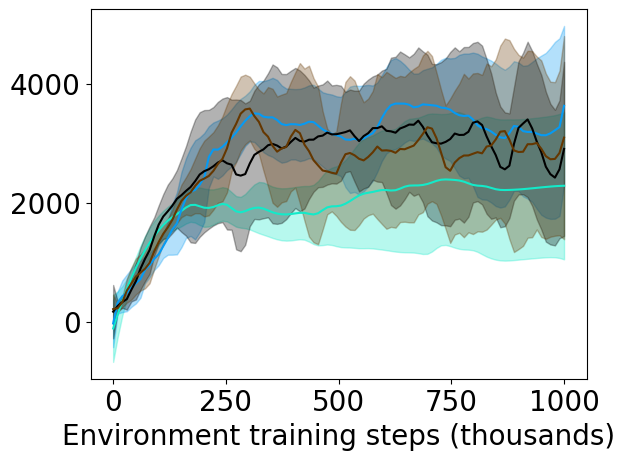}
            \end{subfigure}
            \begin{subfigure}
                    \centering
                    \includegraphics[scale=0.18]{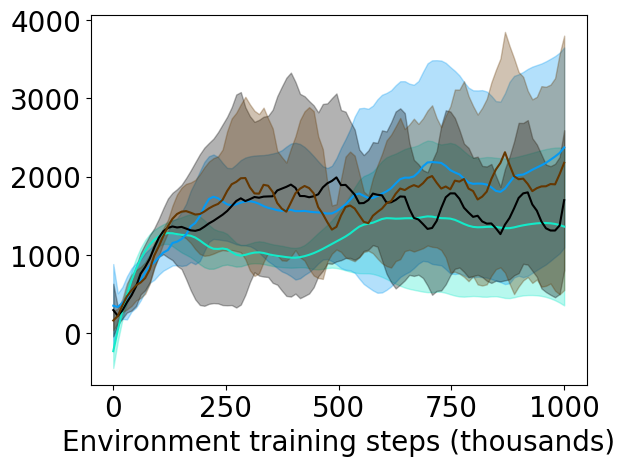}
            \end{subfigure} \\
            \begin{subfigure}
                \centering
                \includegraphics[scale=0.20]{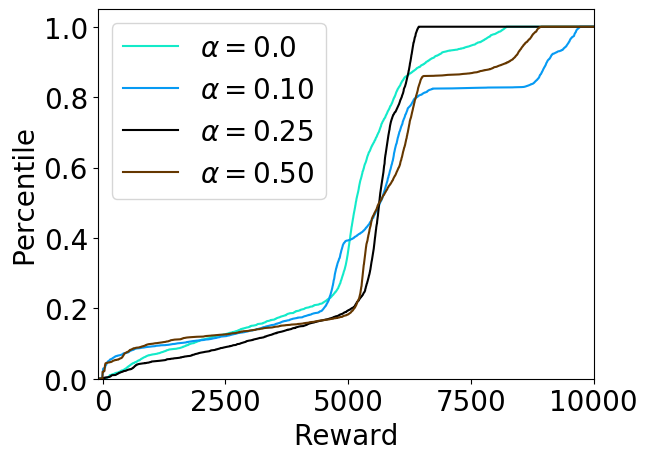}
        \end{subfigure}
        \begin{subfigure}
                \centering
                \includegraphics[scale=0.20]{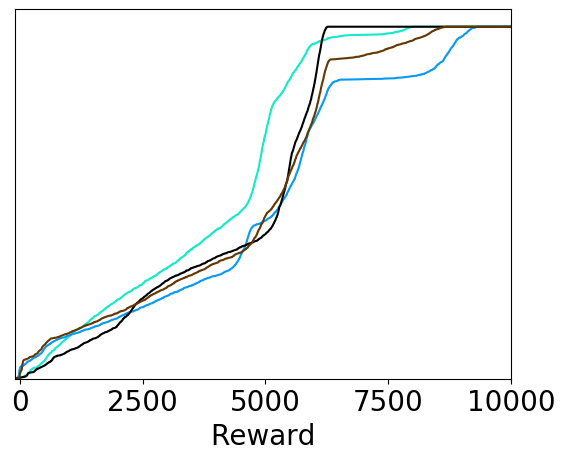}
        \end{subfigure}
        \begin{subfigure}
                \centering
                \includegraphics[scale=0.20]{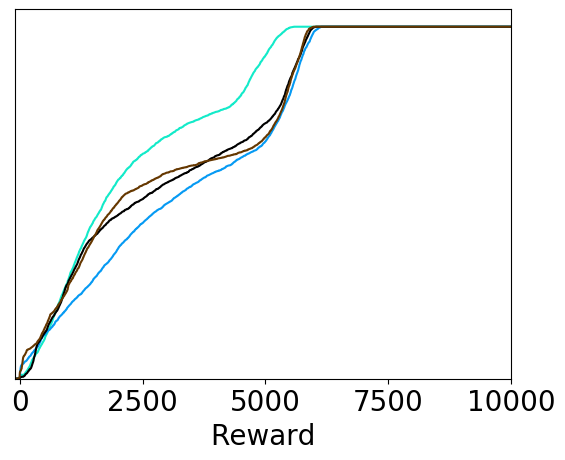}
        \end{subfigure}
        \begin{subfigure}
                \centering
                \includegraphics[scale=0.20]{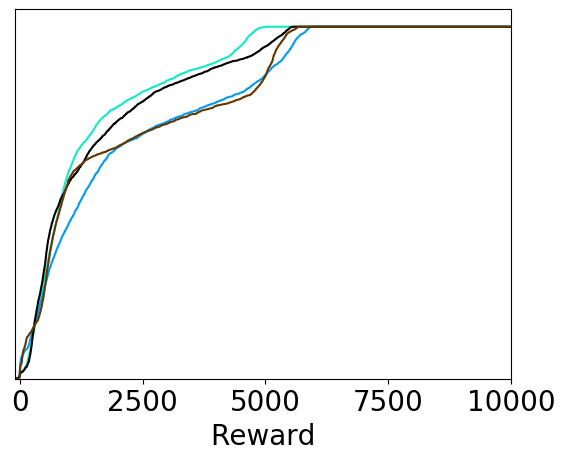}
        \end{subfigure}      
        \vspace{-0.7cm}
            \caption{Walker2d-v2}
    \label{fig:walker2d-v2}
\end{figure*}

Figure \ref{fig:BipedalWalker-v2} shows the performance of our algorithm on BipedalWalker-v2 domain. The top row shows the evaluation curves at different noise levels and the bottom row depicts the corresponding CDFs. It can be observed from the figure that as noise level increases, the performance of all algorithms go down as expected. Moreover, our learned policies with non-zero CVaR $\alpha$ values outperform the risk-neutral one at all the noise levels. 
Figure \ref{fig:HalfCheetah-v2} shows the performance in HalfCheetah-v2 environment. The policy with CVaR value 0.1 show the best performance at all the noise levels. 
Figure \ref{fig:walker2d-v2} shows the evaluation of our algorithm in Walker2d-v2 environment. The risk-neutral one is the most sensitive to the presence of noise. Risk-averse policy with CVaR value 0.5 show the most robustness in either noise free or noise environments. 


\section{Conclusion}
\label{sec_conclusion}
In this paper, we proposed a robust RL algorithm for real-world applications with continuous state action spaces. Our algorithm is based on distributional reinforcement learning which is an idea framework for integrating risk. We utilized sample based policy gradients in this framework and incorporated CVaR to learn risk-averse policies. We demonstrated the robustness of the resulting policies against a range of disturbances in multiple environments. Even though we focused on a special type of risk measure, CVaR, in this work, our framework is compatible with any utility function based risk measure. We will explore these options thoroughly via experiments in the future.

\bibliography{l4dc_2020}

\end{document}